\documentclass[acmtog,nonacm]{acmart}
\acmSubmissionID{1539}

\usepackage{cleveref}
\usepackage{multirow}
\usepackage{listings}
\usepackage[dvipsnames]{xcolor}
\usepackage{subcaption}

\newif\ifshowrevised
\showrevisedfalse 

\ifshowrevised
  \newcommand{\revised}[1]{\textcolor{magenta}{#1}}
\else
  \newcommand{\revised}[1]{#1}
\fi

\usepackage[ruled]{algorithm2e} 

\SetAlFnt{\small}
\SetAlCapFnt{\small}
\SetAlCapNameFnt{\small}
\SetAlCapHSkip{0pt}

\acmJournal{TOG}

\definecolor{yangcolor}{RGB}{255,0,0} 
\definecolor{mikacolor}{RGB}{0,128,0} 
\definecolor{jiaqicolor}{RGB}{0,0,255} 

\begin{document}
\title[Img2CAD: Reverse Engineering 3D CAD Models from Images]{Img2CAD: Reverse Engineering 3D CAD Models from Images through VLM-Assisted Conditional Factorization}

\author{Yang You}
\affiliation{%
 \institution{Stanford University}
 \city{Stanford}
 \state{CA}
 \country{USA}}
\email{yangyou@stanford.edu}

\author{Mikaela Angelina Uy}
\affiliation{
\institution{Stanford University}
 \city{Stanford}
 \state{CA}
 \country{USA}
}
\affiliation{%
  \institution{NVIDIA}
  \city{Santa Clara}
  \state{CA}
  \country{USA}
}
\email{mikacuy@cs.stanford.edu}

\author{Jiaqi Han}
\affiliation{
\institution{Stanford University}
 \city{Stanford}
 \state{CA}
 \country{USA}
}
\email{jiaqihan@stanford.edu}

\author{Rahul Thomas}
\affiliation{
\institution{Stanford University}
 \city{Stanford}
 \state{CA}
 \country{USA}
}
\email{rt03mas@stanford.edu}

\author{Haotong Zhang}
\affiliation{
\institution{Peking University}
 \city{Beijing}
 \country{China}
}
\email{2200012835@stu.pku.edu.cn}

\author{Yi Du}
\affiliation{%
 \institution{Stanford University}
 \city{Stanford}
 \state{CA}
 \country{USA}}
\email{duyi@stanford.edu}

\author{Hansheng Chen}
\affiliation{%
 \institution{Stanford University}
 \city{Stanford}
 \state{CA}
 \country{USA}}
\email{hshchen@stanford.edu}

\author{Francis Engelmann}
\affiliation{%
 \institution{Stanford University}
 \city{Stanford}
 \state{CA}
 \country{USA}}
\email{francis.engelmann@stanford.edu}

\author{Suya You}
\affiliation{
\institution{DEVCOM Army Research Laboratory}
 \city{Los Angeles}
 \country{USA}
}
\email{suya.you.civ@army.mil}

\author{Leonidas Guibas}
\affiliation{%
 \institution{Stanford University}
  \city{Stanford}
 \state{CA}
 \country{USA}
}
\email{guibas@cs.stanford.edu}

\renewcommand{\shortauthors}{You, Y. et al}

\begin{teaserfigure}
  \includegraphics[width=\textwidth]{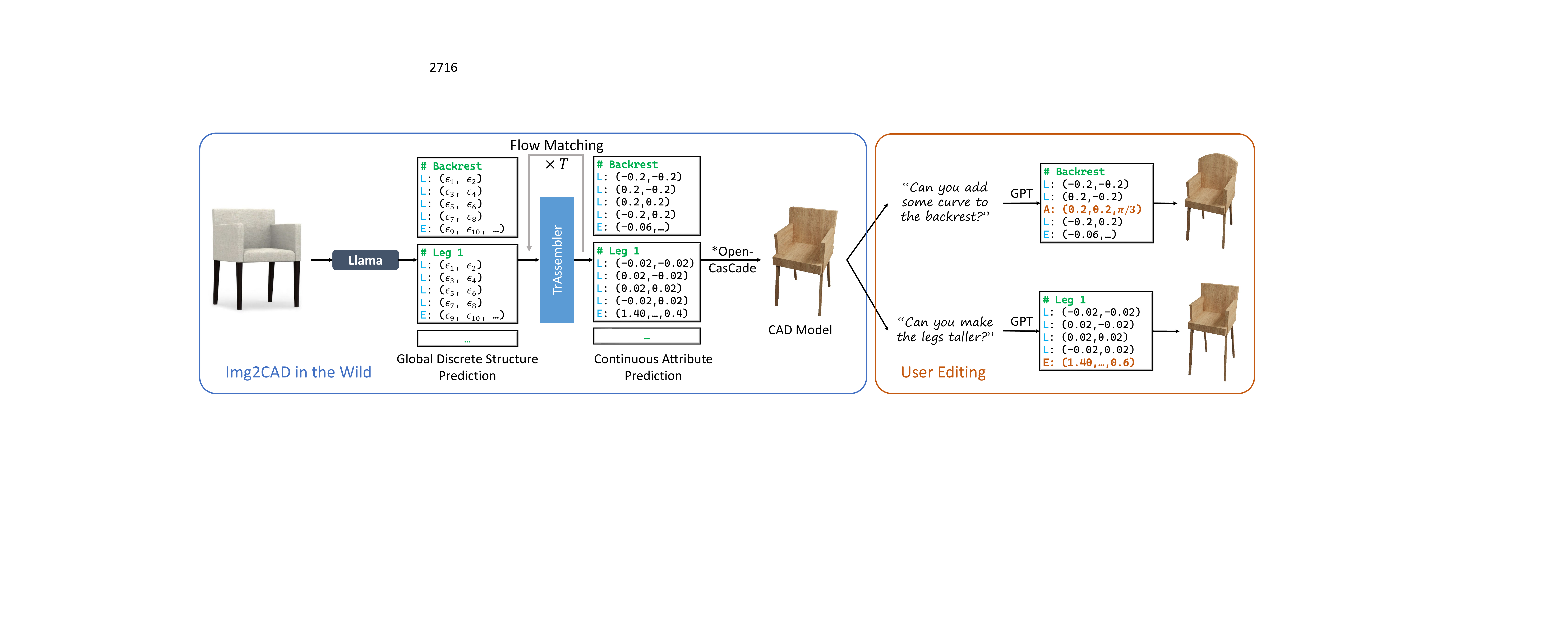}
  \caption{\textbf{Img2CAD: a framework for reverse engineering 3D CAD models from single-view images.} Our method leverages VLM to predict the discrete CAD program structure and then uses a semantic-conditioned transformer to predict the continuous attributes. This approach allows users to easily reconstruct and edit a CAD model from a single-view input image. We use OpenCasCade~\cite{opencascade} to convert the CAD program back to a 3D mesh.}
  \label{fig:teaser}
    \Description{Overview of Img2CAD framework. A single input chair image is processed by a vision-language model and a Transformer to predict CAD program structure and continuous attributes, assembled into a CAD model that can be reconstructed and edited, with examples of modifying the backrest and leg height.}
\end{teaserfigure}

%
%


%
%

\begin{abstract}
Reverse engineering 3D computer-aided design (CAD) models from images is an important task for many downstream applications including interactive editing, manufacturing, architecture, robotics, etc. The difficulty of the task lies in vast representational disparities between the CAD output and the image input. CAD models are precise, programmatic constructs that involves sequential operations combining discrete command structure with continuous attributes -- making it challenging to learn and optimize in an end-to-end fashion. 
Concurrently, input images introduce inherent challenges such as photometric variability and sensor noise, complicating the reverse engineering process. In this work, we introduce a novel approach that conditionally factorizes the task into two sub-problems. First, we leverage vision-language foundation models (VLMs), a finetuned Llama3.2, to predict the global \emph{discrete base structure} with semantic information. Second, we propose TrAssembler that conditioned on the discrete structure with semantics predicts the \emph{continuous attribute} values.
To support the training of our TrAssembler, we further constructed an annotated CAD dataset of common objects from ShapeNet. Putting all together, our approach and data demonstrate significant first steps towards CAD-ifying images in the wild. \revised{Code and data can be found in \href{https://github.com/qq456cvb/Img2CAD}{https://github.com/qq456cvb/Img2CAD}}.
\end{abstract}

\maketitle


\section{Introduction}

A vast array of human-made objects in our everyday environments are created using computer-aided design (CAD)~\cite{VARADY1997255}. Reverse engineering the structural CAD representation~\cite{reverse_engineering_modeling_methods} from raw sensor input is an essential task that enables the fabrication of 3D shapes and their manipulation or editing for various downstream applications, such as manufacturing, design, and robotics. The beauty of a CAD representation lies in its elegant structure, which combines \textit{discrete} operation types with \textit{continuous} attribute parameters. This results in a programmatic, parameterized structure that allows for easy and intuitive editing and manipulation.

Previous works~\cite{uy2022point2cyl, li2023secad, ren2022extrudenet} have explored reverse engineering 3D CAD models from input point clouds, which require 3D sensor inputs. However, images are much more common for everyday objects, such as those captured with smartphones or found in product images via Google search. This makes the task of reverse engineering CAD models from images particularly important and useful. The image-to-CAD problem presents significant challenges, as we typically only obtain a single view of the object in the form of pixels, while the CAD representation demands a detailed understanding of the object's whole 3D geometry.

Reverse engineering CAD models from input images poses two significant challenges: \emph{generalization} and \emph{representation complexity.}
The first challenge lies in the ability of the models to generalize across diverse images, particularly when these images originate from heterogeneous sources. Images from different sources will give different viewing angles, lighting conditions, sensor noise and textures, which makes training a generalizable model from image directly to CAD extremely difficult. This problem is even more severe as there is a lack of CAD data available for common objects in everyday life (e.g., chairs and tables).

Secondly, CAD systems, despite their utility in applications requiring detailed editing and manipulation, present significant challenges due to its combination of discrete command types with continuous parameter values. A CAD model is a sequence of discrete geometric operations of different types, e.g. line, arc, circle, extrude add, extrude cut -- each characterized by continuous attributes such as length, position, and angle. Learning this dual nature of both \emph{discrete} operations and their \emph{continuous} attributes poses a non-trivial challenge, necessitating algorithms to predict both the structural and attribute components accurately.
The hybrid \emph{discrete-continuous} nature of the problem makes image-to-CAD learning non-trivial. 
This may require an exponential amount of data to learn a combinatorial space of structures, which is difficult to obtain.
To tackle these problems, we propose to conditionally factorize the task into two sub-problems -- to first predict the global \emph{discrete base structure} and then, \emph{conditioned} on this base structure, to regress the \emph{continuous attribute values}. We leverage the generalizability of large vision foundation models, particularly a finetuned Llama3.2~\cite{grattafiori2024llama}, to predict the global base structure from an image. We then propose a transformer-based network named \textbf{TrAssembler}, to regress the continuous attributes conditioned on the discreate structure. 
While the space of possible object structures is vast, even within a single category such as chairs, shared substructures among objects are common. For example, many chairs have support structures consisting of four legs, and chairs with headrests are more commonly office chairs with five swivel legs.
Moreover, we also find that VLM's can generate semantically consistent part labels as program comments for such common substructures. These labels can then be used by the continuous attribution prediction network to effectively establish correspondences that facilitate shared learning across \emph{corresponding \revised{attributes}} (e.g., leg length) of different objects, thus enabling the sharing of information across the attribute space.
Our conceptual idea is inspired by the mathematical notion of sheaves~\cite{curry2014sheaves}, where fibers (the attribute parameter spaces in our case) over points in a base space (the space of labelled CAD structures in our case) have partial locally consistent structure (through \emph{corresponding attributes}). \revised{Taking this mathematical concept as inspiration, we leverage shared attributes across different shapes by using shared part names allowing TrAssembler to capture consistency effectively with limited training data.}

To further enhance the quality of the generated CAD models, we incorporate recent flow-matching models, specifically GMFlow~\cite{chen2025gaussian} into our \textbf{TrAssembler}, and employ a masked flow matching loss during training. Given that many human-made objects exhibit strong symmetry priors, we also introduce a symmetry guidance loss during the ODE-based inference sampling. This loss promotes symmetric outputs by performing gradient descent on the estimated score function over the symmetry manifold.

Lastly, to train TrAssembler, we further introduce a CAD dataset of common objects. This was constructed
by leveraging the rich part annotations of PartNet~\cite{mo2019partnet} and the general assistance of GPT4V~\cite{achiam2023gpt} -- resulting in our CAD-ified subset of ShapeNet~\cite{chang2015shapenet}. We will release this dataset upon publication. Our experiments demonstrate the feasibility of our approach for image-to-CAD, which, to the best of our knowledge, is the first of its kind for a general sketch-extrude CAD representation of common objects. We show that our method outperforms baselines and further showcase various applications, such as CAD-ification and shape editing from images in the wild.

\section{Related Works}

\subsection{Reverse Engineering CAD}
Reverse engineering has long been a core topic in graphics and CAD~\cite{BENIERE20131382, VARADY1997255, segmentation_methods_for_smooth_point_regions, Benk2002ConstrainedFI, Ma2023MultiCADCR}. Early methods use inverse CSG modeling~\cite{inversecsg, csgnet, ucsgnet, ren2021csgstump, yu2023d2csg}, but suffer from overly complex decompositions. Recent works leverage large-scale datasets~\cite{koch2019abc, willis2021fusion, Ari2020} to enable learning-based methods for B-Rep analysis~\cite{jayaraman2021uvnet, brepnet, 10044468}, surface inference~\cite{GuoComplexGen2022, liu2023point2cad}, and CAD generation~\revised{\cite{khan2024cadsignet, Lambourne_2022, xu2023hierarchical, liu_sig24, chen2025cadcrafter, li2025caddreamer}}. Unlike these mechanical-part-focused datasets, we introduce a dataset of everyday objects with semantic parts.

The \emph{sketch-extrude} paradigm~\cite{willis2021fusion, uy2022point2cyl, xu2022skexgen, para2021sketchgen} supports compact, flexible primitives and has inspired sequential modeling approaches~\cite{wu2021deepcad, ganin2021computeraided, jayaraman2023solidgen, 10.1093/jcde/qwae005}. Point2Cyl~\cite{uy2022point2cyl} frames reverse engineering as geometry-aware decomposition, extended by later works to unsupervised~\cite{ren2022extrudenet}, self-supervised~\cite{li2023secad}, and multi-view settings~\cite{hong2024mvcyl}. Our method tackles the harder single-image case.

\subsection{Structured Representation of Shapes}
Our daily objects are often highly structured in their geometry and part composition. For example, a chair is composed of parts governed by a set of rules and constraints~\cite{ritchie2023neurosymbolic, uy-deformawareretrieval-eccv20, uy-joint-cvpr21}, such as having legs of equal lengths placed symmetrically around the seat. Existing works have explored various ways of representing and learning structure from shapes, such as part-based templates~\cite{GanapathiSubramanian2018ParsingGU, Kim:2013:LPT, maks_structureDiscovery_sigg11}, symmetries~\cite{mgp_approx_symm_sig_06, Sung:2015}, shape programs~\cite{10.1145/3596711.3596738, jones2020shapeAssembly, jones2021shapeMOD, jones2023ShapeCoder}, and grammars~\cite{10.1145/2185520.2185551, ckgk-prabm-11}. Notably, GRASS~\cite{li_sig17} and StructureNet~\cite{mo2019structurenet} represent a shape as a graph of part hierarchies learned through a supervised neural network. The goal of this line of work is to learn higher-level semantic structures, i.e., parts of shapes, but does not focus \revised{on} the reconstruction of the local geometric detail of each part. These methods typically assume a coarse and homogeneous representation for each part, commonly as bounding boxes. In contrast, our goal is to address image-to-CAD for daily objects, which requires not only understanding the underlying structure but also reconstructing each part's geometry.


\subsection{Primitive Fitting}
Primitive fitting in graphics focuses on decomposing shapes into surface or volumetric primitives. Classical methods detect basic primitives like planes~\cite{czerniawski20186d, fang2018planar, monszpart2015rapter} using RANSAC~\cite{Schnabel2007, li2011globfit}, region growing~\cite{oesau2016planar}, or Hough voting~\cite{borrmann20113d, drost2015local}. More recent data-driven approaches~\cite{spfn, sharma2020parsenet, Le_2021_ICCV, liu_sig24} fit primitives such as planes, cylinders, and B-Reps from point clouds using neural networks. 

Beyond fitting, primitive-based shape abstraction has also been explored using cuboids~\cite{zou20173d, abstractionTulsiani17} or superquadrics~\cite{Paschalidou2019CVPR, Paschalidou2020CVPR}. \revised{A recent work~\cite{jones2024TemplatePrograms} also explored using template programs that models shapes as domain-specific programs and represents primitives as cuboids.} These methods generally assume 3D point cloud inputs and/or rely on a fixed primitive vocabulary. In contrast, we represent shapes using a CAD program with general primitives defined by arbitrary closed loops, enabling a broader class of objects.

Primitive-based decompositions also support shape editing~\cite{spfn} and manipulation~\cite{Sung:2020}. ReparamCAD~\cite{Kodnongbua2023ReparamCADZC} learns a variation space from a single model but requires a box-based program as input. Our approach instead predicts a CAD program directly from a single image without assuming any pre-defined primitives.
\section{Background: Sketch-Extrude CAD}
Before diving into the specifics of our method, we first provide some background and define the general \emph{sketch-extrude} primitives that we use to represent shapes in this work. We adopt a popular industrial standard for reverse engineering shapes into a sequence of sketch and extrude commands, as used in solid modeling softwares such as OpenCascade~\cite{opencascade}, Fusion360~\cite{fusion360}, and Solidworks~\cite{solidworks2005solidworks}. These tools enable the creation of a wide array of shapes, as they do not assume a finite set of primitives but can represent any geometry that is a composition of primitives created by the \emph{extrusion} of any non-self-intersecting, closed \emph{sketch} loop. Below, we define the different types of discrete commands in this CAD language and their corresponding continuous attributes.


\subsection{Sketch.} A sketch is defined as any closed, non-self-intersecting loop drawn on a 2D plane, where each loop is called a profile. A profile is created with a sequence of sketch commands of three different types -- \textit{lines} ($L$), \textit{arcs} ($A$), and \textit{circles} ($R$). The continuous attributes for each type of sketch command are as follows:
\begin{itemize}
    \item $L : (x, y)$, where $(x, y)$ defines the endpoint of a line segment (modellers typically assume that the starting point of the first line segment is at the origin).
    \item $A: (x, y, \alpha)$ where $(x, y)$ is the endpoint of an arc with a sweep angle $\alpha$.
    \item $R: (x, y, r)$, which defines a circle with the center $(x, y)$ and radius $r$. 
\end{itemize}
To result in a valid sketch profile, the resulting curve from the series of commands must be closed and non-self-intersecting. Without loss of generality, we require sketches to be constructed in a counter-clockwise direction.


\begin{figure}[t]
\centering
\includegraphics[width=\linewidth]{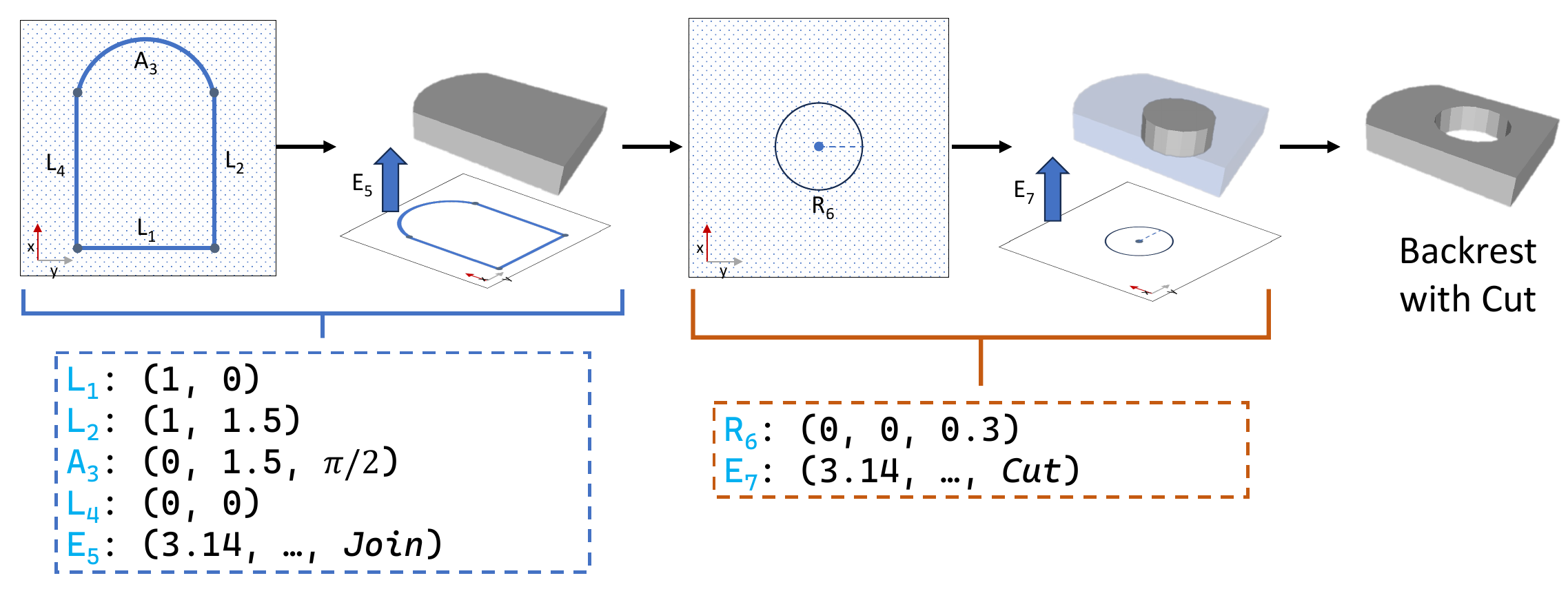}
\vspace{-2em}
\caption{\textbf{An illustrative example of a sketch and extrusion command sequence for a chair's backrest with a circular cutout.} The backrest is created using two extrusion blocks: the first one employs the sketch commands \textit{lines} and \textit{arcs} with the extrusion type \textit{NewBody}, while the second one uses the sketch command \textit{circles} with the extrusion type \textit{Cut}.}
\label{fig:sketchextrude}
\Description{Example of sketch and extrusion sequence for a chair backrest. The backrest is first extruded from a sketch of lines and arcs to form the base shape, then a circular sketch is extruded with the cut operation to create a hole in the backrest.}
\vspace{-1em}
\end{figure}

\subsection{Extrusion.}  In manufacturing, extrusion is the process of pushing sketches forward along a fixed cross-sectional profile to a desired distance, creating a volumetric solid, often referred to as an extrusion cylinder~\cite{uy2022point2cyl}. Extrusion cylinders are then combined through either join or cut operations to form the final model. We distinguish between an extrusion \emph{join} and an extrusion \emph{cut} as two distinct command types in this work.

The continuous attributes of the extrusion command are given as $E:(\alpha, \theta, \gamma, x, y, z, e)$, where $(\alpha, \theta, \gamma)$ are the three Euler angles defining the extrusion coordinate frame, $(x, y, z)$ denote the origin of the extrusion coordinate system, and $e$ represents the extrusion distance or extent. Note that in our implementation, there is an additional $b$ argument specifying whether it is a join or cut. As defined, we regard this as two different command types.




Figure~\ref{fig:sketchextrude}  shows an illustrative example of the sketch-extrude CAD language, comprising discrete command types and their continuous attribute values. The discrete command types define the structure of the underlying shape -- in this case the backrest -- while the attributes continuously vary the shape's geometry. For example, changing the third argument in the $R$ command will enlarge the cut.

\section{Method: Img2CAD}

\begin{figure}[t!]
\centering
\vspace{-1em}
\includegraphics[width=\linewidth]{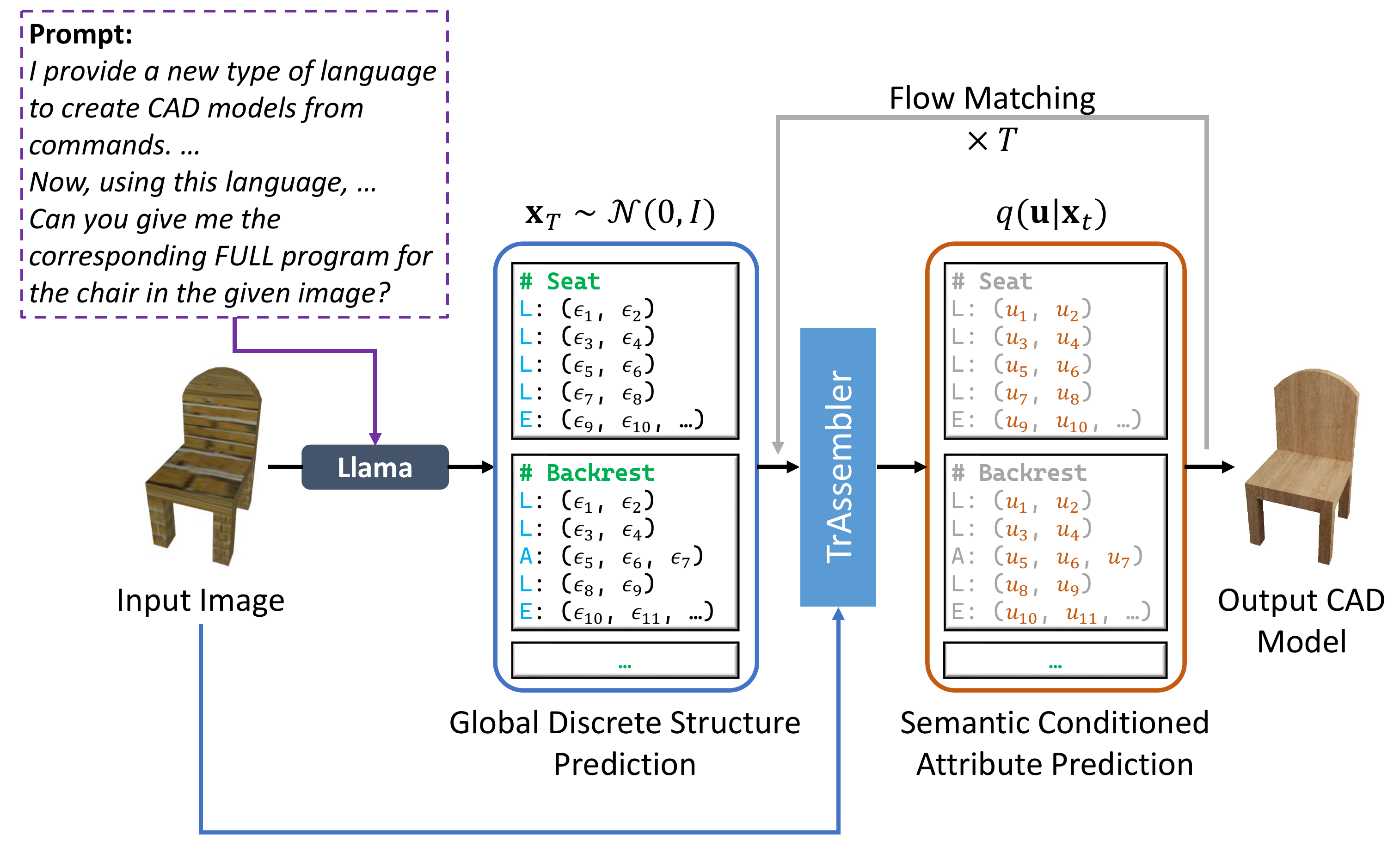}
\vspace{-2em}
\caption{\textbf{Overview of our two-stage pipeline.} The first stage involves Llama3.2 decomposing the input image into semantic parts and generate a sequence of discrete CAD structures. The second stage uses a transformer flow matching model to predict continuous attributes for each part.}
\label{fig:pipeline}
\Description{Diagram of the two-stage Img2CAD pipeline. An input chair image is processed by Llama to generate discrete CAD structures for parts such as the seat and backrest, followed by a flow matching transformer that predicts continuous attributes, producing a complete CAD model.}
\vspace{-1em}
\end{figure}

Reverse engineering a 3D model from input images is a challenging task in computer graphics and vision due to the discrete-continuous nature of the problem. This task requires learning the combinatorial space of shape and program structure along with their corresponding continuous attributes.
Our approach addresses this problem for common objects by \emph{conditionally factorizing} the task into two sub-problems. First, we leverage the capabilities of VLMs (i.e., Llama3.2~\cite{grattafiori2024llama}) to predict the global \emph{discrete base structure} of the shape from a single input image. This discrete problem involves inferring the \emph{semantic} parts present in the image and providing the \emph{CAD structure}, i.e., the command types of each underlying part. Subsequently, we propose a novel transformer-based network called \textbf{TrAssembler} that, conditioned on this semantic discrete base structure, predicts the \emph{continuous attributes} for the sequence of CAD commands for each semantic part, with flow matching loss and inference time symmetry guidance. An overview of the pipeline is shown in Figure~\ref{fig:pipeline} and we elaborate on each component in the following subsections.

\subsection{Discrete Global Structure with VLM}
\label{sec:met_discrete}

In this work, we define the global \emph{discrete base structure} of a shape as its part decomposition along with the corresponding CAD command types for each constituent part. 
Inferring a shape’s discrete base structure from a single image is inherently challenging due to the combinatorial complexity and the difficulty of generalizing across diverse input images. 
To address this, we leverage VLMs, which have been trained on large-scale data. Furthermore, we observe that VLMs such as LLama3.2 or even GPT-4o~\cite{achiam2023gpt} may sometimes struggle to produce accurate structural predictions, primarily due to their limited inherent understanding of 3D geometry. Hence, we finetune an open-sourced VLM with our training data to achieve better results. Concretely, our VLM uses LLama3.2 finetuned on predicting the discrete base structure of the shapes from our data, i.e. it is not trained on and does not predict the continuous attributes which we leave to our continuous attribute network.

\begin{figure}[t!]
\centering
\includegraphics[width=0.95\linewidth]{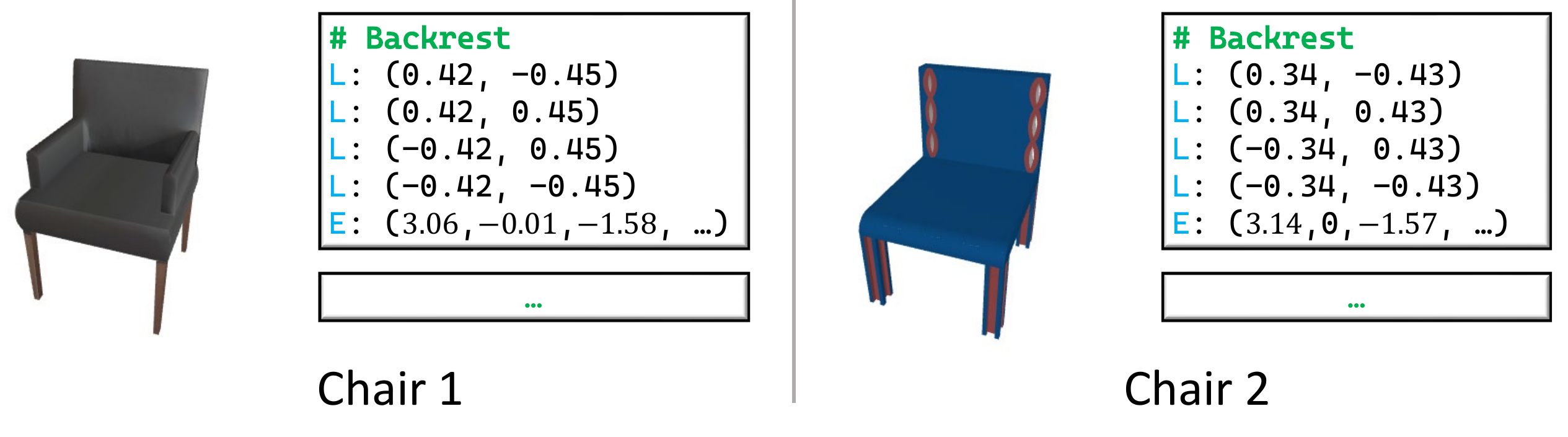}
\vspace{-1em}
\caption{Different chairs still share similar part semantics, and similar parts share similar attribute parameters, facilitating easier attribute prediction.}
\vspace{-1em}
\label{fig:gptprompt}
\Description{Comparison of two chairs showing that both have backrests represented with similar semantic parts and attribute parameters, illustrating consistency across different chair designs.}
\end{figure}

We prompt the VLM to provide \emph{semantic part comments} in its predicted global base structure. This involves attaching a part label, before providing the CAD sequence for each identified part. 
This semantic labeling is crucial as it facilitates the shared learning across corresponding attributes of different objects in our attribute prediction network. This design allows for sharing between shapes with common substructures, even when they have different global discrete base structures. For example, as shown in Figure~\ref{fig:gptprompt}, backrests from different chairs, recognized and parsed by Llama3.2, are represented within distinct semantic structural blocks, yet the ground-truth CAD commands for these blocks share similar parameters such as axis of extrusion, dimensions, and extrusion distances.

\begin{figure*}[ht]
\centering
\includegraphics[width=0.9\linewidth]{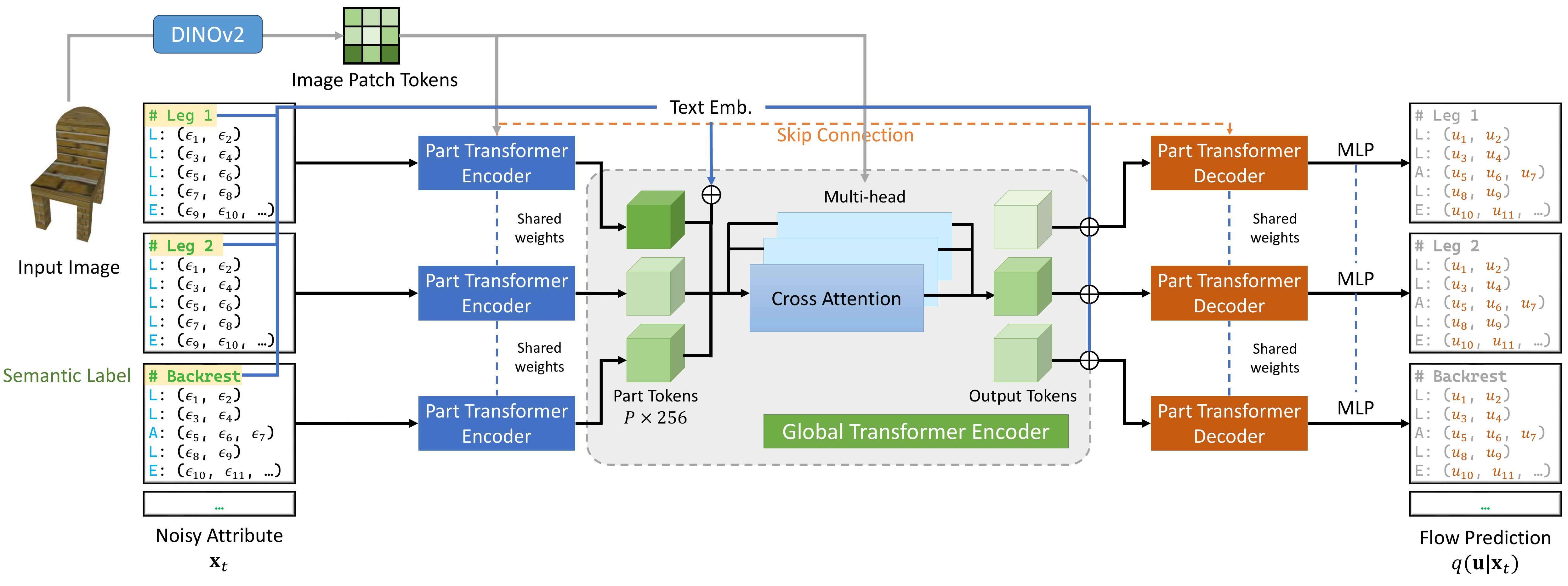}
\vspace{-1.5em}
\caption{\textbf{Overall architecture of TrAssembler.} The model first leverages part transformer encoder to generate part embeddings with part structures given by Llama. The part embeddings are then fed into a global transformer encoder and refines them with multi-head self-attention, followed by an MLP to decode the final attribute parameters for each command.}
\label{fig:network}
\Description{Architecture of TrAssembler. Part transformer encoders process semantic parts of the input chair image, the outputs are refined through a global transformer encoder with cross-attention, and part decoders predict continuous attributes for each part.}
\vspace{-0.5em}
\end{figure*}

\subsection{TrAssembler -- Continuous Attribute Prediction in Shared Space}

\begin{figure}[ht]
\centering
\includegraphics[width=\linewidth]{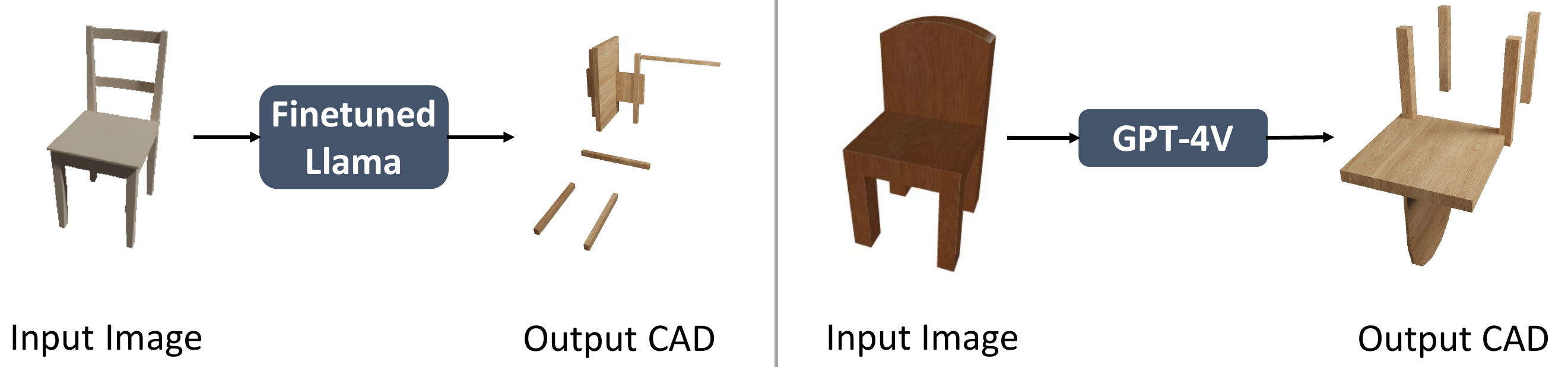}
\vspace{-2em}
\caption{\textbf{Illustration of VLM's limitations in predicting continuous attributes.} While VLMs effectively decomposes labeled parts, it struggles with predicting accurate continuous attributes.}
\vspace{-1em}
\label{fig:gptfail}
\Description{Examples showing limitations of vision-language models in CAD reconstruction. Input chair images are correctly decomposed into labeled parts, but the output CAD models have incorrect continuous attributes, leading to misaligned or distorted parts.}
\end{figure}

Our exploration shows that despite VLM being good at inferring the discrete base structure, part semantics and their underlying sequence of command types, it exhibits limitations in accurately predicting the continuous attributes of CAD commands. Figure~\ref{fig:gptfail} illustrates a failure case of VLMs, even with Llama3.2 finetuned on \emph{both} structure and continuous attributes. To address this, we introduce TrAssembler, which takes in a single image and conditioned on the discrete base structure from our finetuned VLM (\ref{sec:met_discrete}), uses flow matching to denoise noisy \textit{continuous attributes} $\mathbf{x}_t$ for all CAD commands. $t$ is the denoising timestep.


TrAssembler first encodes each part by processing its command sequence and noisy attribute inputs through a dedicated part transformer, incorporating cross-attention with DINOv2 image features to produce part-level tokens. These tokens are \revised{concatenated} with CLIP-based semantic embeddings and globally refined via a transformer to capture inter-part relationships. Finally, a decoder predicts denoising flows for each command attribute by cross-attending to learnable parameter tokens and regressing flow scalars via an MLP. The overall architecture of TrAssembler is illustrated in Figure~\ref{fig:network}. 

To sample continuous CAD attributes, we employ GMFlow~\cite{chen2025gaussian} as our flow matching generative model~\cite{flowmatching,liu2022flow}, which predicts a dynamic Gaussian mixture distribution $q(\mathbf{u} | \mathbf{x}_t)$ of flow velocity $\mathbf{u}$ conditioned on noisy attributes $\mathbf{x}_t$ at diffusion time $t$. The model is trained with the negative log-likelihood loss on flow velocity:
\begin{equation}
    \mathcal{L} = \mathbb{E}_{t, \mathbf{x}_0, \mathbf{x}_t} \left[ -\log q\left( \frac{\mathbf{x}_t - \mathbf{x}_0}{t} \middle| \mathbf{x}_t \right) \right],
\end{equation}
where $\mathbf{x}_0$, $\mathbf{x}_t$ are ground truth and noisy attributes, respectively. 
Compared to standard flow matching, GMFlow better captures multi-modal characteristics of CAD attributes. 
Please see supplementary for more details on our net architecture and GMFlow.

A core innovation of TrAssembler lies in its utilization of the discrete structure provided by VLM. By conditioning on this discrete structure, the continuous attribute prediction becomes significantly easier,
as the image-to-CAD task conditioned on the discrete base structure simplifies into a continuous problem. Also critical to our TrAssembler is its hierarchical transformer design, comprising the part and global transformers, and its use of consistent semantic labels from VLM. These elements facilitate information sharing across the attribute space, which our experiments show is essential for achieving better performance.
Additionally, our design choice of using transformer-based architectures enables the handling of varying numbers of parts and different numbers of commands.

\subsection{Inference Time Symmetry Guidance}
Many human-made objects exhibit strong symmetry, a property often overlooked by standard flow-matching or diffusion methods. To address this, we introduce symmetry score guidance during each inference ODE solver step of GMFlow. After each inference step, we refine $\mathbf{x}_t$ by performing gradient descent through our estimated score function over the symmetric data manifold:
\begin{equation}
\mathbf{x}_t = \mathbf{x}_t - \lambda \nabla\log p_{\text{sym}}(t) = \mathbf{x}_t - \lambda \frac{\partial \mathcal{L}_{\text{sym}}}{\partial \mathbf{x}_t},
\end{equation}
where $\mathcal{L}_{\text{sym}}$ denotes the symmetry loss. This loss is computed by sampling a set of points, reflecting them across a symmetry plane or rotating them according to a detected axis of rotational symmetry, and then measuring the absolute differences between the signed distance functions (SDFs) of the original and transformed sets. This guidance enforces symmetry manifold constraint on the CAD attributes $\mathbf{x}_t$.
Additionally, we prompt and query GPT-4o to identify the specific type of symmetry present in the input image, e.g., planar or rotational symmetry. In this work, we optimize for a single symmetry type per shape, which we find to be sufficient for producing consistent improvements.

\begin{table*}[t!]
  \centering
  \small
  \caption{\textbf{Image to CAD reconstruction results comparison on CAD-ified Chair, Table, and Cabinet test dataset.}}
  \vspace{-1em}
 \resizebox{\linewidth}{!}{
 \begin{tabular}{lccccccccc}
    \toprule
          & \multicolumn{3}{c}{Chair} & \multicolumn{3}{c}{Table} & \multicolumn{3}{c}{Cabinet} \\
          \cmidrule(lr){2-4}\cmidrule(lr){5-7}\cmidrule(lr){8-10}
          & CD $\downarrow$    & Seg Acc (\%) $\uparrow$ & Seg mIoU (\%) $\uparrow$ & CD $\downarrow$   & Seg Acc (\%) $\uparrow$ & Seg mIoU (\%) $\uparrow$ & CD $\downarrow$   & Seg Acc (\%) $\uparrow$ & Seg mIoU (\%) $\uparrow$ \\
    \midrule
    GPT-4o & 0.3806 & 50.49 & 34.96 & 0.3676 & 62.16 & 41.99 & 0.3807 & 54.66 & 47.65 \\
    Image-3D-CAD &  0.3266 & 49.03 & 40.73   &  0.3828 & 33.72 & 28.34     &  0.4775 & 59.61 & 49.80 \\
    DeepCAD & 0.2914 & 70.44 & 54.86 & 0.3633 & 60.20  & 46.50  & 0.3310 & 62.70  & 54.95 \\
    DeepCAD-End2End & 0.2346 & 77.01 & 65.06 & 0.3698 & 67.37 & 52.21 & 0.3279 & 60.55 & 53.23 \\
    \midrule
    Ours & \textbf{0.0984} & \textbf{91.86} & \textbf{87.88} & \textbf{0.0966} & \textbf{93.81} & \textbf{82.13} & \textbf{0.1573} & \textbf{73.08} & \textbf{57.59} \\
    \bottomrule
    \end{tabular}%
    }
  \label{tab:main_results}%
  \vspace{-1em}
\end{table*}%

\begin{figure}[t!]
    \centering
    \includegraphics[width=0.9\linewidth]{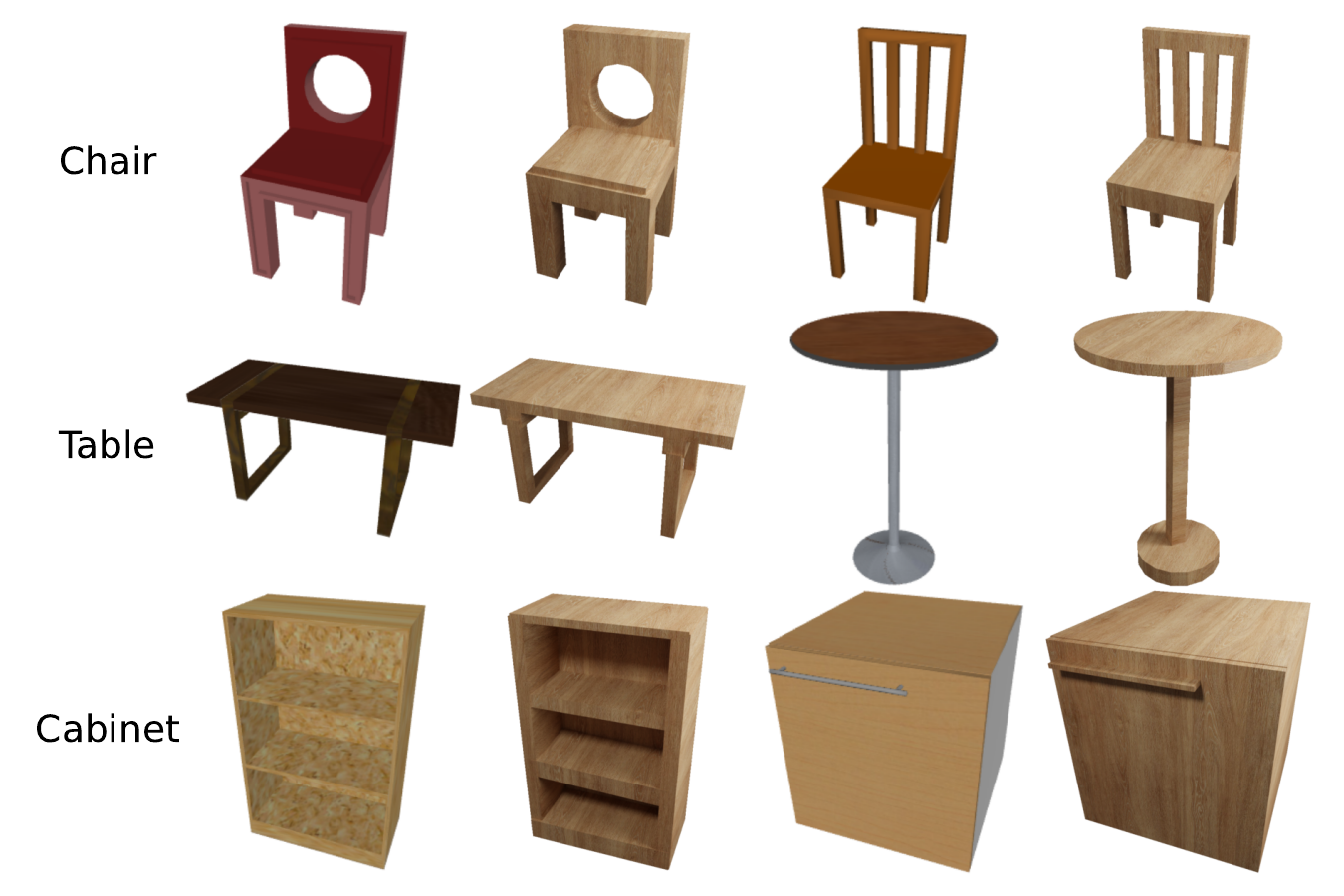}
    \vspace{-1em}
    \caption{\textbf{Data samples from our curated CAD dataset.} For each pair, \textbf{left:} ShapeNet model, \textbf{right:} our CAD-ified model.}
    \label{fig:dataset}
    \Description{Examples of dataset samples showing three categories: chairs, tables, and cabinets. Each pair compares a ShapeNet model on the left with its CAD-converted counterpart on the right.}
    \vspace{-2em}
\end{figure}

\section{Dataset}
\label{sec:dataset}
We need to train TrAssembler with the ground-truth attribute parameters for each part. Therefore, a substantial, curated CAD dataset is necessary. Previous resources such as the ABC dataset~\cite{koch2019abc} offer extensive CAD data; however, these datasets predominantly focus on industrial components and are less applicable to domestic items like chairs and tables. To address this gap and further facilitate CAD research on everyday objects, we have compiled a CAD-ified dataset of common objects encompassing three classes from ShapeNet~\cite{chang2015shapenet}: \textit{chair}, \textit{table}, and \textit{cabinet}. \revised{We select these categories because their corresponding part semantic labels are annotated in PartNet~\cite{mo2019partnet}, which are subsequently used in the dataset creation process}.

For dataset creation, we employ a hybrid method that capitalizes on GPT-4V's structural priors in a semi-automatic fashion. \revised{First, since we have the ground-truth segmentation}, we input the rendered images with segmented parts \revised{that we highlight in red (see supplementary for an example)} into GPT-4V, which then predicts the semantic label and the corresponding lines, arcs, and circles for each part. Subsequently, utilizing the ground-truth \revised{PartNet} labels, we perform one-to-one part label matching using the linear sum assignment method~\cite{burkard2012assignment}. For each matched part prediction, we adjust the \revised{predicted} 3D mesh to correct the scale and translation errors introduced by GPT-4V's assembly inaccuracies. \revised{The resulting adjusted 3D meshes serve as the ground truth labels in our curated dataset. More details can be found in our supplementary.} 


After data generation, each sample undergoes manual human review to ensure its quality \revised{about which 30\% are kept)}, culminating in a comprehensive dataset comprising 1,026 chairs, 3,243 tables, and 305 cabinets. We follow PartNet~\cite{mo2019partnet} train/val/test splits. To the best of our knowledge, this is the first large-scale dataset for common domestic objects, and it is intended to serve as a benchmark for training and evaluating CAD-related tasks on common objects.


\section{Experiments}

In this section, we evaluate our proposed method named \textbf{Img2CAD} for reverse engineering 3D CAD models from images of various common objects. Additionally, we conduct ablation studies to verify the efficacy of our key design elements in the pipeline. Lastly, we show that our method is robust to images from arbitrary viewpoints.

\begin{table*}[t!]
  \centering
    \setlength{\tabcolsep}{7.5pt}
  \small
  \caption{\textbf{Ablation study on TrAssembler.}}
  \vspace{-1em}
  \resizebox{\linewidth}{!}{
    \begin{tabular}{lccccccccc}
    \toprule
          & \multicolumn{3}{c}{Chair} & \multicolumn{3}{c}{Table} & \multicolumn{3}{c}{Cabinet} \\
          \cmidrule(lr){2-4}\cmidrule(lr){5-7}\cmidrule(lr){8-10}
          & CD $\downarrow$   & Seg Acc $\uparrow$ & Seg mIoU $\uparrow$ & CD $\downarrow$   & Seg Acc $\uparrow$ & Seg mIoU $\uparrow$ & CD $\downarrow$   & Seg Acc $\uparrow$ & Seg mIoU $\uparrow$ \\
    \midrule
    (i)\ \ \ \ Llama3.2 end-to-end & 0.2568 & 62.98 & 50.12 & 0.2432 & 72.87 & 61.49 & 0.3707 & 59.18 & 55.11 \\
    (ii)\ \ \ Our finetuned VLM + Flat TrAssembler & 0.3282 & 64.64 & 49.90 & 0.2670 & 76.53 & 56.07 & 0.3013 & 62.99 & 54.77 \\
    (iii) \ Our finetuned VLM + Hierarchical Design & 0.2847 & 63.14 & 44.06 & 0.2645 & 49.10 & 32.17 & 0.2947 & 59.78 & 50.21 \\
    (iv)\ \ iii + Semantic Part Info & 0.1685 & 81.79 & 73.32 & 0.1655 & 83.75 & 69.55 & 0.2327 & 66.32 & 57.33 \\
    (v)\ \ \ iv + Flow Matching & \underline{0.0992} & \underline{91.45} & \underline{86.82} & \underline{0.0993} & \textbf{94.33} & \textbf{82.49} & \textbf{0.1554} & \underline{71.81} & \textbf{59.03} \\
    (vi)\ \ v + Symmetry Guidance (Ours) & \textbf{0.0984} & \textbf{91.86} & \textbf{87.88} & \textbf{0.0966} & \underline{93.81} & \underline{82.13} & \underline{0.1573} & \textbf{73.08} & \underline{57.59} \\
    \bottomrule
    \end{tabular}%
    }
  \label{tab:main_ablations}%
  \vspace{-1em}
\end{table*}%


\subsection{Baselines.} We consider the following four baselines:
\begin{itemize}
\item \emph{GPT-4o}: This baseline directly employs GPT-4o to give the CAD program of the input.

\item \emph{Image-3D-CAD}: Since most CAD-related works require a 3D mesh as input, we build this baseline to test how well current CAD reconstruction methods perform when converting a 2D image to 3D. We first use LRM~\cite{hong2023lrm} to predict the corresponding mesh of the input image. 

Then, the mesh is sent to a self-supervised SECAD-Net~\cite{li2023secad} to predict the final CAD program.
\item \emph{DeepCAD}: We load the checkpoint of DeepCAD pretrained on the ABC dataset~\cite{koch2019abc} and finetune it on our dataset. To enable image-to-CAD reconstruction, we follow the original approach in \cite{wu2021deepcad}, which involves a second-stage training to align the image encoder with the program encoder by minimizing the L2 loss over the latent vectors while keeping the decoder frozen.
\item \emph{DeepCAD-End2End}: We also include a variant of DeepCAD that jointly finetunes the image encoder and program decoder in an end-to-end manner. 
\end{itemize}

\subsection{Evaluation metrics.} We employ three metrics that cover both geometric and semantic proximity:
\begin{itemize}
\item \emph{Chamfer Distance} (CD): This is computed as the Chamfer distance between two point clouds uniformly sampled on the surface of the reconstructed mesh from the model-predicted CAD program and the ground truth.
\item \emph{Part Segmentation Accuracy} (Seg Acc \%): This measures how semantically close the reconstructed object is to the ground truth, specified as the part segmentation accuracy over the point cloud sampled on the surface. We leverage PointNet~\cite{qi2017pointnet} pretrained on ShapeNet~\cite{chang2015shapenet} to produce the segmentation from the generated CAD models. We then compare these segmentations with the ground-truth segmentation labels from ShapeNet to obtain Seg Acc and Seg mIoU. More details in supplementary.
\item \emph{Part Segmentation mIoU} (Seg mIoU \%): This metric also assesses semantic closeness, specified as the mean Intersection over Union (mIoU) of the part segmentation over the point cloud sampled on the surface. 
\end{itemize}


 
\subsection{Main results.} 
\begin{figure*}[ht]
    \includegraphics[width=0.95\textwidth]{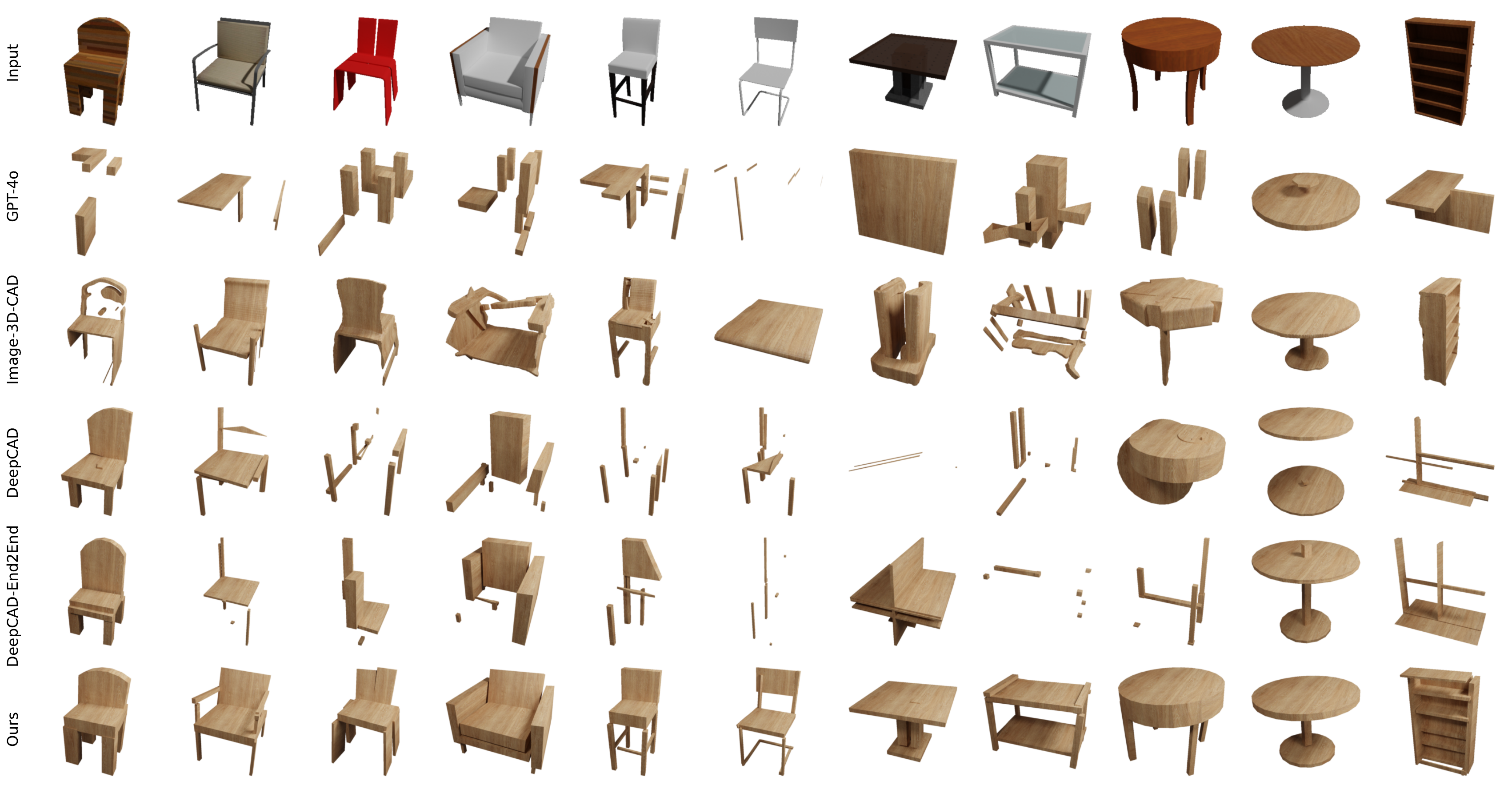}
    \vspace{-1em}
    \caption{\textbf{Qualitative results between different methods on the test split of ShapeNet CADified dataset.} Our method achieves much better results.}
    \label{fig:shapenet}
    \Description{Comparison of CAD reconstruction results on the ShapeNet CADified dataset. Input images are shown on top, followed by outputs from GPT-4o, Image-3D-CAD, DeepCAD, DeepCAD-End2End, and our method at the bottom, which produces more complete and accurate CAD models.}
\end{figure*}

We present the image-to-CAD reconstruction results of our method against the baselines in Table~\ref{tab:main_results}. Qualitative results are illustrated in Figure~\ref{fig:shapenet}. Img2CAD consistently outperforms the baselines across all three object categories and metrics, demonstrating its strong capability in producing both geometrically and structurally accurate CAD models from image inputs. Notably, Img2CAD reduces the average Chamfer distance from 0.3108 to \textbf{0.1174} and improves segmentation accuracy and mIoU by \textbf{17.94\%} and \textbf{19.03\%}, respectively, over the best-performing baseline, DeepCAD-End2End.

GPT-4o performs worse than the other approaches, validating that even state-of-the-art VLMs are insufficient to predict both the program structure and continuous attributes directly. This supports our approach of conditionally factorizing the task: using the program structure predicted by VLMs and offloading the attribute prediction to a learned network. Both DeepCAD and DeepCAD-End2End achieve better performance than GPT-4o but are still markedly inferior to Img2CAD. 

\subsection{Ablation studies.} 
\begin{figure*}[ht]
    \includegraphics[width=0.95\textwidth]{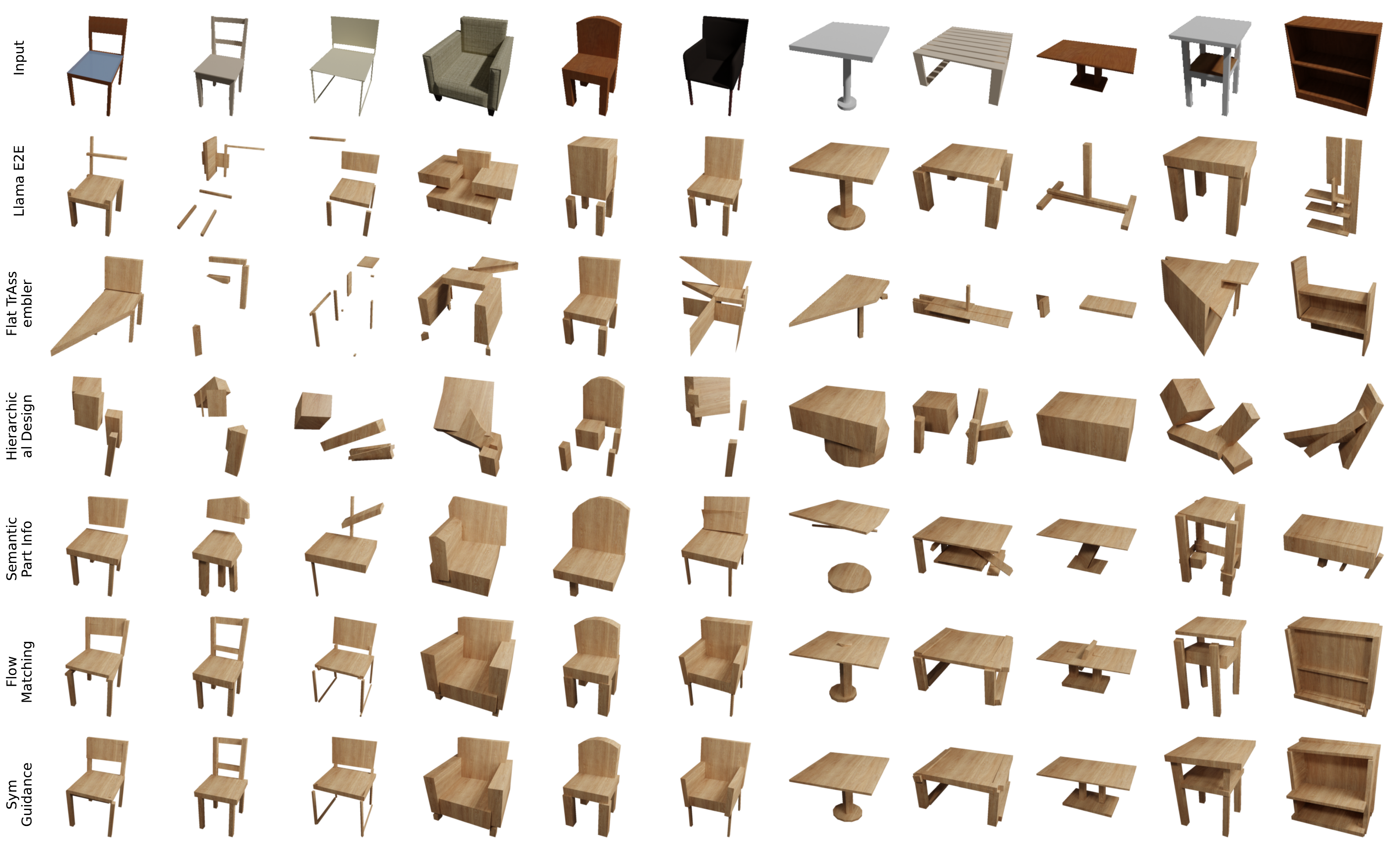}
    \caption{\textbf{Qualitative comparisons of our ablation studies.}}
    \label{fig:ablation}
    \Description{Qualitative results of ablation studies. Input images are shown on top, followed by outputs from different model variants including Llama E2E, Flat-TrAss Assembler, Hierarchical Design, Semantic Part Info, Flow Matching, and Symmetry Guidance, demonstrating how each component contributes to improved CAD reconstruction quality.}
    \vspace{-1em}
\end{figure*}

We ablate several core designs of our TrAssembler with results depicted in Table~\ref{tab:main_ablations} and qualitatively visualized in Figure~\ref{fig:ablation}. Note that these variants are all conditioned on the program structure predicted by Llama.



\begin{itemize}
  \item[(i)] \textbf{Llama3.2 end-to-end:} First we ablate on the necessity of decomposing the task into predicting the discrete global structure and continuous attributes. Specifically, we compare against a variant in which LLaMA3.2 is directly finetuned to predict \emph{both} CAD structures and attributes.
  \item[(ii)] \textbf{Flat TrAssembler:} We apply conditional factorization by using our finetuned VLM on discrete global structure, but use a flat transformer that treats all attributes from different parts as a single 1D token sequence. These are processed through a non-hierarchical transformer.
  \item[(iii)] \textbf{Hierarchical Design:} We apply conditional factorization by using our finetuned VLM on discrete global structure, and further introduce our hierarchical transformer architecture, consisting of the part transformer and the global transformer, without part label information.
  \item[(iv)] \textbf{Semantic Part Info:} We then further incorporate semantic information by adding part-specific text embeddings, enabling the model to better reason about part semantics.
  \item[(v)] \textbf{Flow Matching:} We replace the naive feed-forward prediction head with our flow matching-based loss.
  \item[(vi)] \textbf{Symmetry Score Guidance:} At test time, we further introduce a symmetry-aware guidance strategy to encourage more regular and symmetric outputs.
\end{itemize}

Quantitative results are reported in Table~\ref{tab:main_ablations} and qualitative visualizations are shown in Figure~\ref{fig:ablation}. Experimental results indicate that this approach fails to generalize well to test-time shapes, primarily due to LLaMA3.2’s limited capacity for reasoning over 3D geometry. Moreover, the results show that the hierarchical transformer design and semantic part comments together improve performance by capturing both local and global part-level interactions. The flow matching strategy provides further gains through a diffusion-inspired structured prediction. Finally, symmetry guidance at inference time helps refine the output structure, better visualized in Figure~\ref{fig:ablation}.

\subsection{Symmetry and Connectivity.}
To evaluate the structural fidelity of the generated CAD programs, we report two quantitative metrics averaged across all object categories: the number of strongly connected components (SCCs) and the symmetry Chamfer distance. The SCC metric captures geometric disconnections by counting the number of connected components in the output mesh, where two vertices are considered connected if their Euclidean distance is below a threshold of 0.05. Fewer SCCs indicate more coherent and physically plausible structures (fewer floating components). The symmetry Chamfer distance measures how closely the generated shape aligns with its mirror reflection over the X-axis, using bidirectional Chamfer distance between point clouds sampled from the original and mirrored meshes. As shown in Table~\ref{tab:structure_metrics}, introducing hierarchical design with semantic part information already yields notable improvements in both connectivity and symmetry. Our full model, equipped with symmetry guidance, further reduces structural fragmentation and enhances global regularity, producing CAD programs that are both functionally and geometrically well-formed.

\begin{table}[ht]
  \centering
  \small
  \caption{\textbf{Structural quality analysis:} number of strongly connected components (SCC, lower is better) and symmetry chamfer (lower is better).}
  \vspace{-1em}
  \setlength{\tabcolsep}{5pt}
  \resizebox{\linewidth}{!}{
  \begin{tabular}{lcccc}
    \toprule
    & \#SCC $\downarrow$ & SymChamfer $\downarrow$ \\
    \midrule
    (i)\ \ \ \ Llama3.2 end-to-end & 2.06 & 0.1268 \\
    (ii)\ \ \ Our finetuned VLM + Flat TrAssembler & 1.79 & 0.1502  \\
    (iii) \ Our finetuned VLM + Hierarchical Design & 1.90 & 0.2748  \\
    (iv)\ \ iii + Semantic Part Info & 1.49 & 0.1176  \\
    (v)\ \ \ iv + Flow Matching & 1.16 & 0.0827  \\
    (vi)\ \ v + Symmetry Guidance (Ours) & \textbf{1.11} & \textbf{0.0756}  \\
    \bottomrule
  \end{tabular}
  }
  \label{tab:structure_metrics}
\end{table}

\subsection{Robustness on arbitrary viewpoints.}
\begin{figure}[ht]
    \centering
    \vspace{-1em}
    \includegraphics[width=\linewidth]{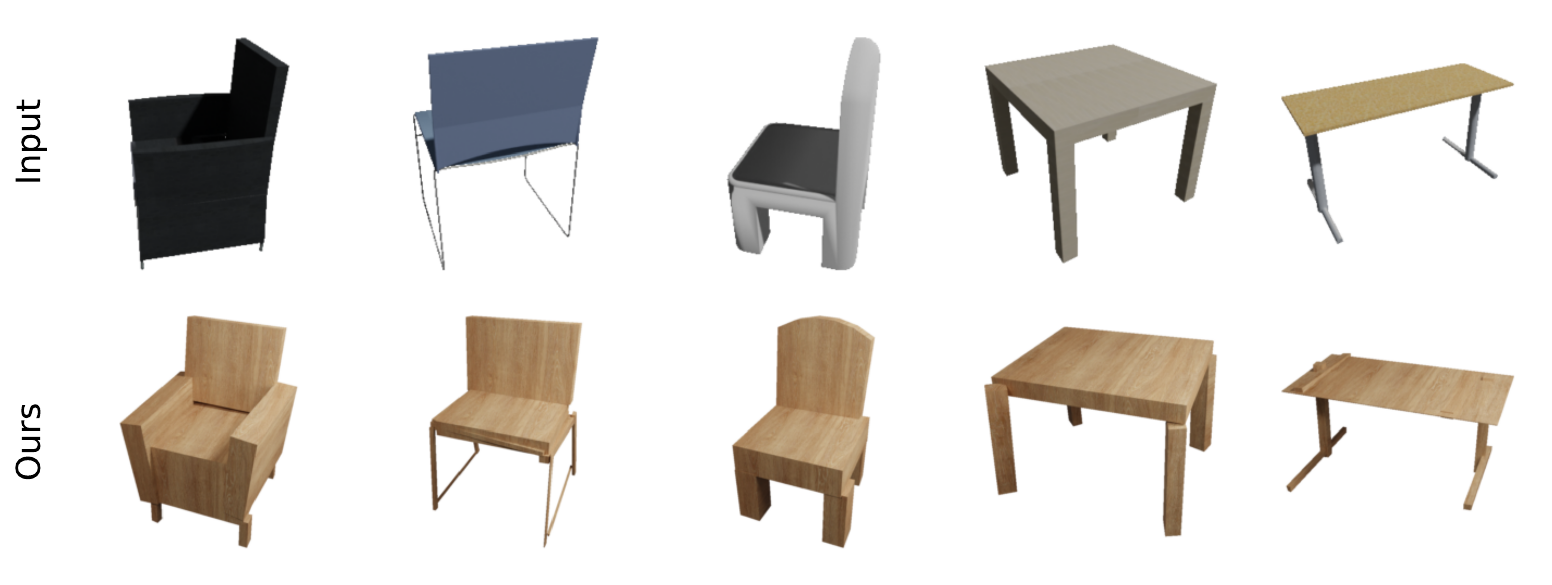}
    \vspace{-1em}
    \caption{\textbf{Example output visualizations given arbitrary-view inputs.}}
    \label{fig:arbview}
    \Description{Examples showing CAD reconstructions from arbitrary-view input images. The top row shows chairs and tables from unusual viewpoints, while the bottom row shows corresponding CAD models reconstructed by our method.}
\end{figure}

\begin{figure}[ht]
\centering
\vspace{-1em}
\includegraphics[width=\linewidth]{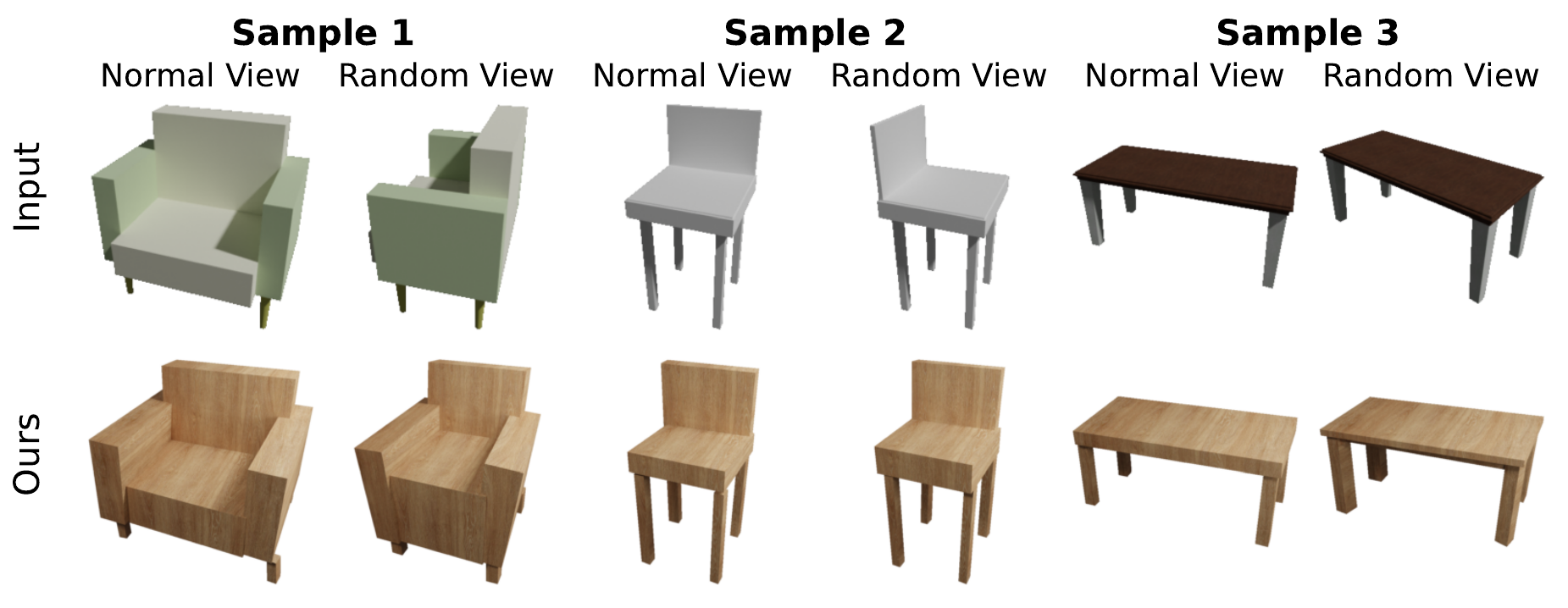}
\caption{\textbf{Example visualizations of reconstruction results from frontal and arbitrary-view inputs.}}
\label{fig:randview_vis}
\Description{Reconstruction examples from both frontal and arbitrary-view inputs. The top row shows input chairs and tables in normal and random viewpoints, while the bottom row shows the corresponding CAD models generated by our method.}
\vspace{-1em}
\end{figure}

\begin{table}[t!]
  \centering
  \small
  \setlength{\tabcolsep}{3.8pt}
  \caption{\textbf{Our model achieves comparable performance on arbitrary views.} Results are averaged over categories.}
  \vspace{-1em}
    \begin{tabular}{lccc}
    \toprule 
     & CD $\downarrow$ & Seg Acc $\uparrow$ & Seg mIoU $\uparrow$\\
    \midrule
    Frontal View & 0.1174 & 86.25 & 75.87 \\
    Arbitrary View & 0.1396 & 84.41 & 73.53 \\
    \bottomrule
    \end{tabular}%
  \label{tab:viewinv}%
\end{table}%
Our method is also very robust thanks to the view invariance provided by VLMs(i.e., Llama, DINO and CLIP). As shown in Table~\ref{tab:viewinv}, we get a slight drop on performance when evaluated on images sampled from arbitrary viewpoints. This important property helps make our method generalizable to images in the wild, even when it is trained on frontal-view images, as shown in Figure~\ref{fig:arbview}. 
\revised{We also computed the Chamfer Distance (CD) between the instance-wise CAD reconstructions from frontal and arbitrary views, obtaining a low CD of 0.0282. Qualitative comparisons between the two views are presented in Figure~\ref{fig:randview_vis}.}

\section{Applications}

\subsection{Image-to-CAD in the Wild}

\begin{figure*}[ht]
    \centering
    \includegraphics[width=\textwidth]{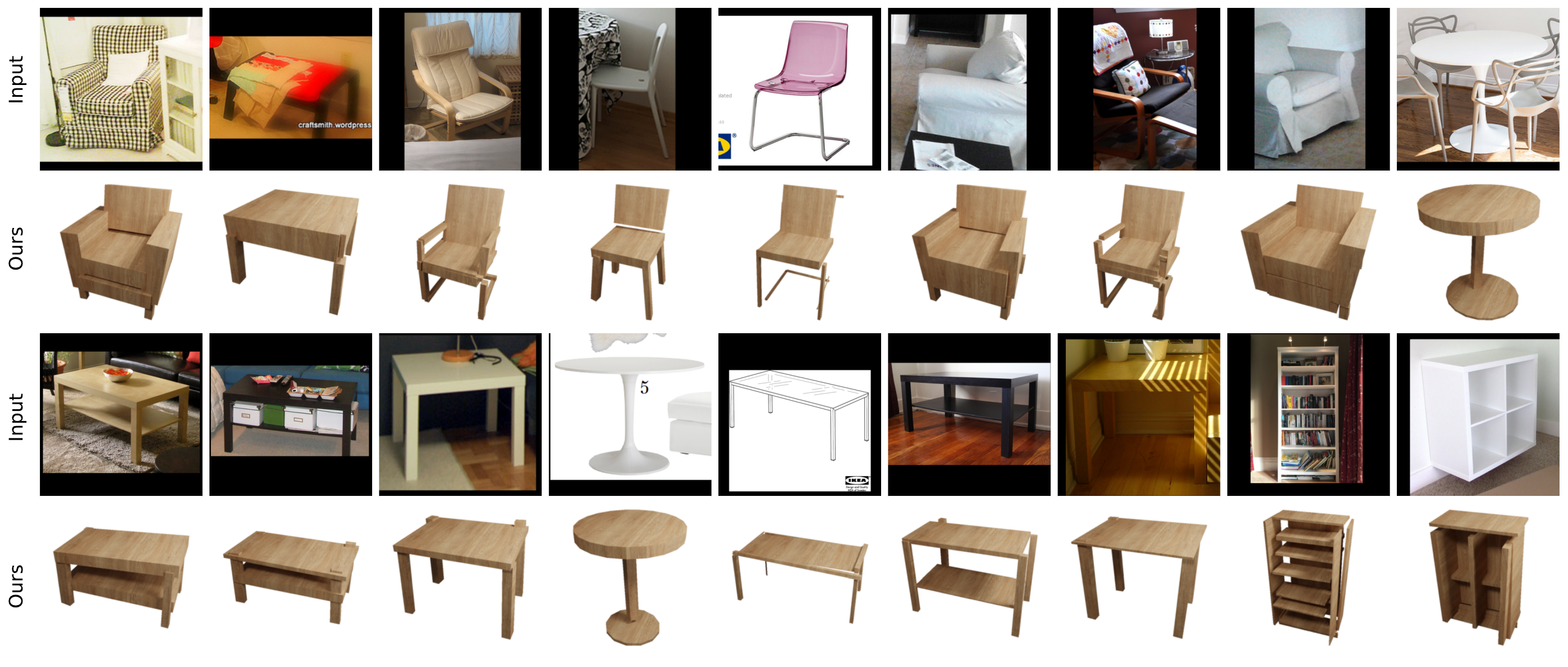}
    \vspace{-1em}
    \caption{\textbf{Image to CAD results on Pix3D dataset.}}
    \label{fig:pix3d}
    \Description{Qualitative results on the Pix3D dataset. The first rows show real-world input images of chairs, tables, and cabinets, while the second rows show the corresponding CAD models reconstructed by our method.}
    \vspace{-1em}
\end{figure*}

\begin{figure*}[htbp]
\centering
\begin{minipage}[t]{0.55\textwidth}
    \centering
    \includegraphics[width=\linewidth]{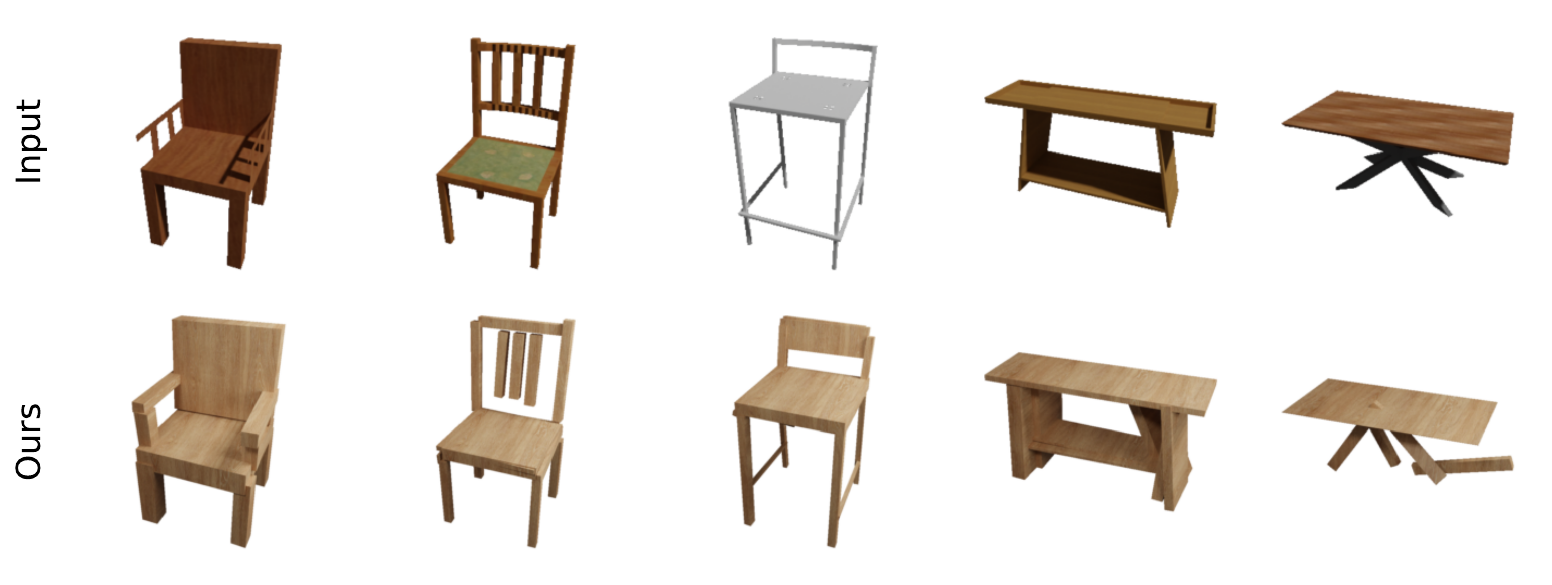}
    \caption{\textbf{Failure cases.} Llama3.2 can sometimes hallucinate or miss some structural parts, and TrAssembler can struggle on predicting the accurate position on complex long-tailed examples (right-most).}
    \label{fig:failure}
    \Description{Failure cases of our method. Input images of chairs and tables are shown on top, with reconstructed CAD models below. Some results miss structural parts or have inaccurate positioning, especially in complex long-tailed examples.}
\end{minipage}
\hfill
\begin{minipage}[t]{0.4\textwidth}
    \centering
    \includegraphics[width=\linewidth]{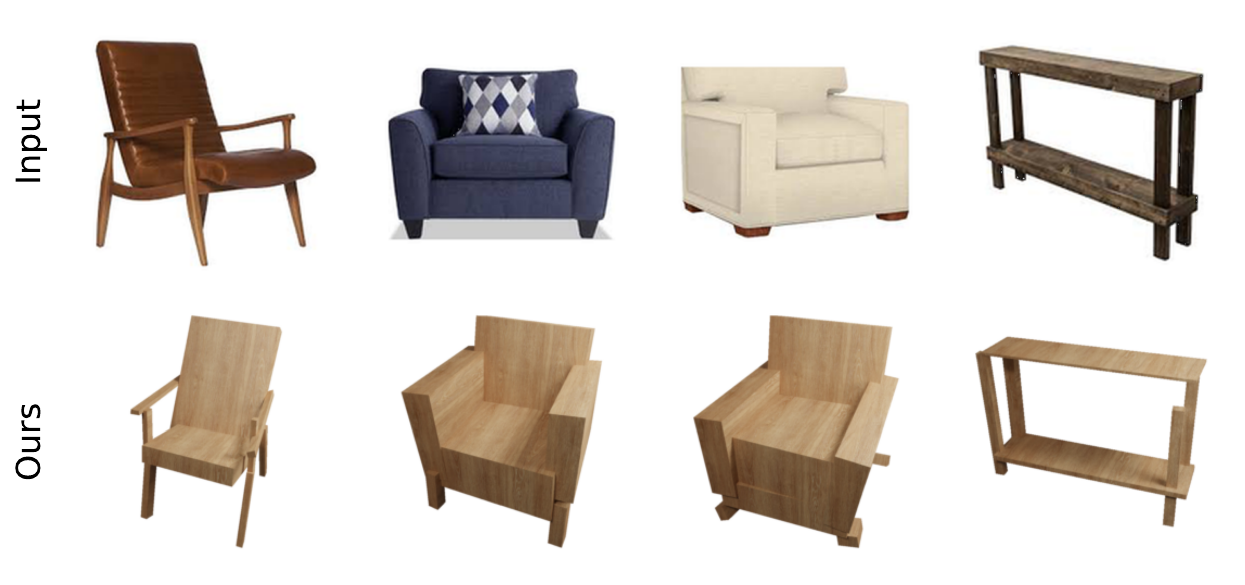}
    \caption{\textbf{Image to CAD results on Google furniture images.}}
    \label{fig:google}
    \Description{Qualitative results on Google furniture images. The top row shows input photos of chairs, sofas, and a table, while the bottom row shows the corresponding CAD reconstructions produced by our method.}
\end{minipage}
\end{figure*}

Although our model is trained exclusively on synthetic ShapeNet CAD-ified models, it demonstrates the ability to generalize to real-world images. This is achieved through the foundational knowledge provided by VLMs. Figures~\ref{fig:pix3d} and~\ref{fig:google} present qualitative results using images from Pix3D~\cite{sun2018pix3d} and Google furniture images.
This task is exceptionally challenging due to the difficulties associated with \textit{generalization} and \textit{CAD representation complexity}.

\begin{figure*}[ht]
    \centering
    \includegraphics[width=\textwidth]{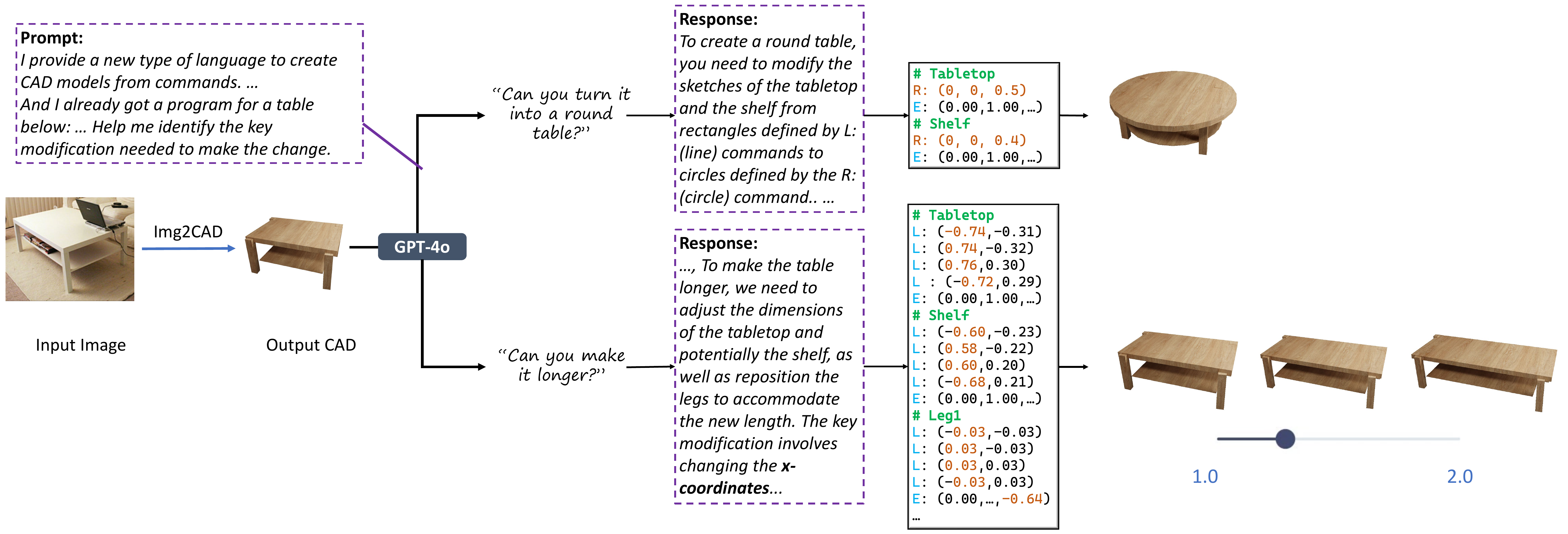}
    \vspace{-1em}
    \caption{\textbf{Image conditioned CAD editing example.} Users can easily create multiple variations of a table from a single-view image.}
    \Description{Image showing multiple variations of a table generated from a single-view image using CAD editing.}
    \label{fig:editing2}
    \vspace{-1em}
\end{figure*}


\subsection{Structure-Aware Editing.}
A key advantage of our approach is its support for structure-aware editing in real-world scenarios, enabled by our interpretable CAD representation. As illustrated in Figure~\ref{fig:editing2}, users can intuitively modify an object's CAD structure within an image using natural language instructions. The process begins by retrieving the CAD program using Img2CAD. This program is then provided to GPT-4o, which identifies the necessary modifications to achieve the user-specified outcome.
It supports both structural changes, e.g., transforming a rectangular table into a round one, and continuous attribute adjustments, such as modifying the table length. GPT-4o exhibits strong reasoning abilities by understanding part relationships: it adjusts both the tabletop and the legs accordingly. 
\section{Conclusion}
We propose a novel approach for reverse engineering 3D CAD models from input images by conditionally factorizing this challenging task into two sub-problems. First, we leverage VLMs (i.e., Llama3.2) to infer the discrete global structure with semantic information. We then introduce TrAssembler, which, conditioned on this discrete global structure, predicts the continuous attribute values of the CAD program. Additionally, we constructed a CAD dataset of common objects from ShapeNet to support the training of the continuous attribute prediction network. Our experiments demonstrate promising results, significantly outperforming  baselines. Our method also makes a significant step towards generating CAD models from images in the wild, showcasing its potential for real-world applications.

\section{Limitations and Future Work.} 
Although flow matching achieves strong results, the sampling time remains dependent on the number of inference timesteps. For instance, using the Euler ODE solver with 32 steps requires approximately 5 seconds to generate a single sample, highlighting a limitation of diffusion-like models. We anticipate that incorporating recent one-step diffusion methods could significantly reduce inference time, which we leave for future work.

\revised{Furthermore, perfect symmetry and connectivity are not guaranteed by our inference-time symmetry guidance. This guidance acts as a soft constraint, as each flow-matching step alternates between a standard flow update (which remains on the learned manifold) and a symmetry-score update (which encourages symmetry). This interleaving biases the CAD parameters toward symmetry while maintaining validity, resulting in improved symmetry and connectivity in most cases (Figure~\ref{fig:shapenet}, Table~\ref{tab:main_ablations}), if not always. In future work, we plan to investigate methods for enforcing symmetry and connectivity directly within the flow-matching steps.}

Despite finetuning, Llama3.2 remains imperfect. As shown in Figure~\ref{fig:failure}, the predicted structure occasionally contains missing or hallucinated parts, indicating challenges in handling complex or long-tailed cases. To quantify the potential upper bound, we conducted an experiment where the predicted structure from Llama3.2 was replaced with the ground-truth CAD structure. This substitution led to a reduction in Chamfer Distance (CD) from 0.1174 to 0.1032. Future work could focus on enhancing the robustness of language models in predicting accurate and complete CAD programs.




\begin{acks}
This work is supported by ARL grant W911NF-21-2-0104, a Vannevar Bush Faculty Fellowship, and a gift from the Flexiv corporation. Yang You also acknowledges the support from the Outstanding Doctoral Graduates Development Scholarship of Shanghai Jiao Tong University.
\end{acks}

\newpage

\bibliographystyle{ACM-Reference-Format}
\bibliography{main}


\end{document}